# Application of machine learning algorithm in temperature field reconstruction

A preprint

Qianyu He
Huazhong University of Science and Technology,m202371763@hust.edu.cn

Huaiwei Sun
Huazhong University of Science and Technology,hsun@hust.edu.cn

**Yubo Li**
Huazhong University of Science and Technology,d202480719@hust.edu.cn

**Zhiwen You**
Huazhong University of Science and Technology,m202271591@hust.edu.cn

**Qiming Zheng**
The Chinese University of Hong Kong
qmzheng@cuhk.edu.hk

**Yinghan Huang**
Zhongnan Engineering Corporation Limited, 1220075860@qq.com

**Sipeng Zhu**
Huazhong University of Science and Technology,zsp960512@163.com

**Fengyu Wang**
Yunnan Hexu Environmental Technology Co., Ltd.,wangfy@hexutech.cn

February, 2025

## Abstract

This study focuses on the stratification patterns and dynamic evolution of reservoir water temperatures, aiming to estimate and reconstruct the temperature field using limited and noisy local measurement data. Due to complex measurement environments and technical limitations, obtaining complete temperature information for reservoirs is highly challenging. Therefore, accurately reconstructing the temperature field from a small number of local data points has become a critical scientific issue. To address this, the study employs Proper Orthogonal Decomposition (POD) and sparse representation methods to reconstruct the temperature field based on temperature data from a limited number of local measurement points. The results indicate that satisfactory reconstruction can be achieved when the number of POD basis functions is set to 2 and the number of measurement points is 10. Under different water intake depths, the reconstruction errors of both POD and sparse representation methods remain stable at around 0.15, fully validating the effectiveness of these methods in reconstructing the temperature field based on limited local temperature data. Additionally, the study further explores the distribution characteristics of reconstruction errors for POD and sparse representation methods under different water level intervals, analyzing the optimal measurement point layout scheme and potential limitations of the reconstruction methods in this case. This research not only effectively reduces measurement costs and computational resource consumption but also provides a new technical approach for reservoir temperature analysis, holding significant theoretical and practical importance.

# 1. Introduction

Thermal stratification is an important characteristic of large water bodies such as oceans, lakes, and reservoirs (Norri et al., 2020), typically manifesting prominently during the warmer periods of the year. This stratification phenomenon primarily occurs due to factors such as air temperature and solar radiation, which cause the upper layer of water to warm up. The increase in temperature reduces the density of the upper layer (Maria et al., 2024), triggering heat exchange and convective mixing. Water temperature stratification leads to significant differences in the aquatic environment at various depths of reservoirs, thereby affecting the distribution and activities of aquatic organisms. Additionally, the discharge of cold water from the reservoir's hypolimnion can have substantial impacts on downstream agriculture and fisheries (Zhang et al., 2023). Therefore, studying the phenomenon of water temperature stratification not only helps optimize the functionality of existing reservoirs and enhance their benefits but also promotes more efficient utilization of water resources, holding significant importance for ecological environmental protection.

With the continuous improvement in the scale and resolution of modern experimental methods and numerical simulations, a wealth of fluid flow data has been generated. Although unprecedented measurement and simulation fidelity can be achieved in laboratory environments, practical applications are often limited to a few noisy sensors. Thus, the challenge of flow field estimation lies in integrating large amounts of offline data with limited and unreliable online information. In recent years, the use of Proper Orthogonal Decomposition (POD) dimensionality reduction methods for temperature distribution reconstruction and analysis has yielded promising results, becoming a research hotspot (Gao et al,2019). Gappy POD was introduced to repair damaged or missing data (Xing et al,2022) and has been successfully applied to reconstruct unsteady flow fields around airfoils (Peng et al,2023) and low-dimensional ocean velocity and temperature fields (Yildrim et al., 2009). However, these methods are mostly based on least squares regression, which may be prone to overfitting and sensitivity to noise. Moreover, although these methods minimize the kinetic energy deviation between predictions and actual measurements, this does not guarantee that the reconstruction will be globally optimal.

Sparse representation has made significant progress in regularization for inverse problems (such as denoising and inpainting) as well as in computer vision and pattern recognition (Song et al,2021).Sparse representation has demonstrated effectiveness in the three-dimensional image reconstruction of planar array ECT (Rebollo-Neira et al., 2019).In magnetic particle imaging, wavelet-based sparse representation methods have also been used for image reconstruction (Nguyen et al., 2023).Additionally, sparse representation can prevent overfitting and provide robustness to noisy and corrupted measurements, which is crucial for flow field estimation.

# 2. Data and methods

## 2.1. Study area

Lake Diefenbaker, an elongated reservoir in western Canada, serves as an exemplary site for studying reservoir hydrology and thermal dynamics. Spanning 181.6 km in length with a maximum width of 6 km, the reservoir has a surface elevation of 556.87 m at full pool level. It covers an area of 393 km², with a maximum depth of 60 m and a total storage capacity of 9.03 km³. The reservoir's average inflow of 254 m³/s and mean residence time of 1.2 years contribute to its distinct thermal stratification patterns, which are crucial for understanding the effects of reservoir dynamics on water temperature. The majority of its inflow (95%) comes from the South Saskatchewan River, while the remaining 5% is supplied by Swift Current Creek and other minor tributaries. Since its commissioning in 1967, Lake Diefenbaker has fulfilled multiple roles, including hydropower generation (275 MW), agricultural irrigation (supporting 70% of Saskatchewan's irrigated lands), municipal-industrial water supply, and flood control. In recent years, it has also provided ecosystem services such as commercial aquaculture, recreational tourism, and biodiversity conservation, supporting 29 fish species and 187 waterfowl species. The reservoir's thermal stratification is influenced by both natural factors, such as inflow rates and seasonal variations, and human interventions, including water level fluctuations driven by hydropower and irrigation demands. Its complex inflow regime, dominated by the South Saskatchewan River with contributions from minor tributaries, creates thermal gradients that make it an ideal site for studying mixed-layer dynamics in large reservoirs. Located at the prairie-boreal ecotone and serving as a key stopover on the Central Flyway, Lake Diefenbaker is highly sensitive to temperature variations. This sensitivity provides valuable insights into how thermal changes in regulated systems impact ecosystem biodiversity, particularly migratory species. Furthermore, its responsiveness to climate change enhances its significance as a model for tracking shifts in reservoir thermal structures.

## 2.2. Data resource

The data utilized in this study were derived from the CE-QUAL-W2 model simulations of Lake Diefenbaker, as documented in the study (Lindenschmidt et al., 2019) is dataset encompasses comprehensive water quality and hydrodynamic parameters, including water temperature, dissolved oxygen, total phosphorus, total nitrogen, and other nutrient dynamics, simulated over the period of 2011–2013. The model scenarios explored the impacts of six different water extraction depths on in-reservoir water quality dynamics, providing a detailed understanding of how dam outflow elevation influences nutrient cycling and biogeochemical processes. This publicly available dataset, obtained from the Federated Research Data Repository (FRDR) [1], offers a robust foundation for investigating the complex interactions between dam operations and reservoir water quality, facilitating further analysis and application in related research.

## 2.3. Sparse representation

Sparse representation is a data-driven framework for reconstructing high-dimensional flow fields from limited and potentially noisy measurements. The core idea is to approximate the target flow field as a sparse linear combination of examples from a preconstructed library, leveraging prior knowledge of coherent structures observed in historical data. This approach enhances robustness against noise and overfitting compared to traditional least-squares methods. The methodology is structured as follows:

A training library $\Psi \in R^{n \times r}$ is built using snapshots of flow fields from simulations or experiments, where $n$ is the dimensionality of the discretized field, and $r$ is the number of training examples. Optionally, the empirical mean flow $\bar{x}$ is subtracted to focus on fluctuating components. The library may include raw training data, tailored modes (e.g., POD modes), or learned dictionaries (e.g., K-SVD).

Measurements $y \in R^p$ are obtained via a linear operator $C \in R^{p \times n}$ applied to the true field $x$:

$$y = Cx + \eta \qquad (1)$$

where $\eta$ represents measurement noise or corruption. Common configurations include point sensors, slices, or downsampled regions.

The flow field xx is approximated as $x \approx \Psi s$, where $s \in R^r$ is a sparse coefficient vector. To recover ss, the following convex optimization problem is solved:

$$\hat{s} = \arg \min_{s} \| s \|_1 \text{ subject to } \| y - C\Psi s \|_2 \leq \epsilon \qquad (2)$$

Here, $\| \cdot \|_1$ promotes sparsity, while the constraint ensures consistency with measurements within a tolerance $\epsilon$, adjusted based on noise level $\sigma$. For gross corruption, an augmented formulation jointly minimizes $\| s \|_1 + \| e \|_1$, where $e$ accounts for sparse outliers.

The full field is reconstructed as:

$$\hat{x} \approx \Psi \hat{s} \qquad (3)$$

Optionally, the amplitude is rescaled to match the energy of training data, particularly for noisy cases where solutions may underestimate fluctuations.

## 2.4. Error Evaluation

In this study, we employed several metrics to evaluate the accuracy of flow field reconstruction using sparse representation methods. The primary metric used was the normalized root-mean-square error (RMSE), which quantifies the difference between the reconstructed flow field $\hat{x}$ and the true flow field $x$. The error was defined as:

$$error_1 = \frac{\|x - \hat{x}\|_2}{\|x\|_2} \tag{4}$$

where $\|\cdot\|_2$ denotes the Euclidean norm. This metric normalizes the reconstruction error by the magnitude of the true field, providing a dimensionless measure of accuracy that is independent of the field's scale.

In cases where the empirical mean $\bar{x}$ was subtracted from both the true and reconstructed fields to highlight fluctuations, the error metric was modified as:

$$error_2 = \frac{\|x - \hat{x}\|_2}{\|x + \bar{x}\|_2} \tag{5}$$

This adjustment ensures that the error calculation focuses on the deviations from the mean rather than the overall field magnitude.

## 3. Results

### 3.1. Reconstruction Performance with Varied POD Bases and Measurement Points

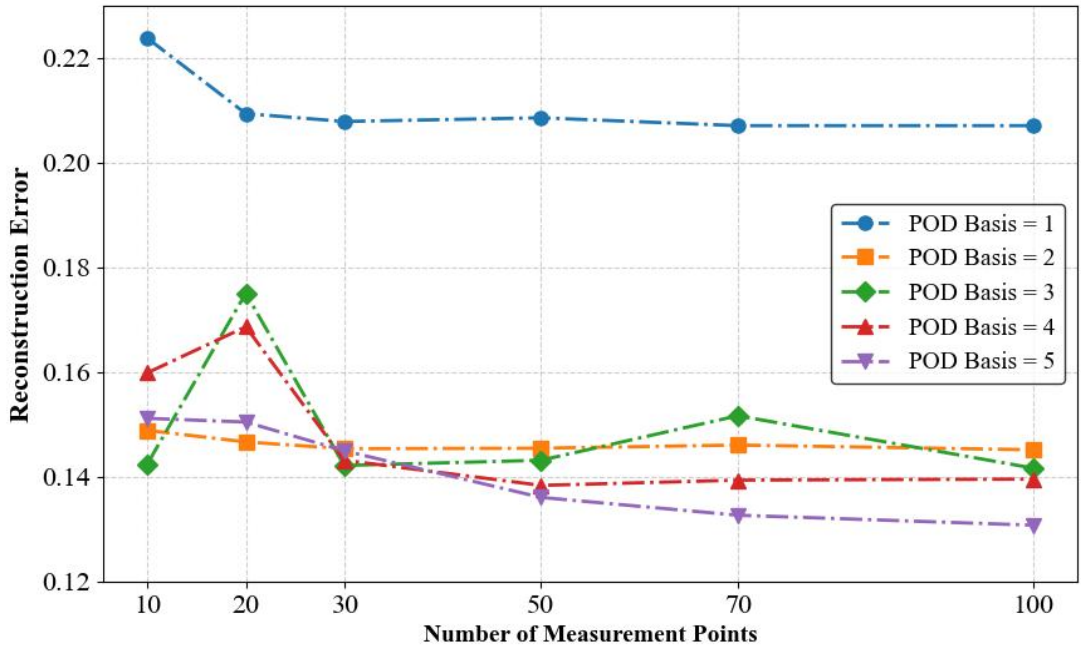

Figure 1 Reconstruction error with varying POD bases and measurement points

To analyze the impact of factors such as the number of POD bases, the number of sensors (measurement points), and the placement of sensors (measurement locations) on the reconstruction accuracy of the temperature field, a controlled variable approach was employed. To mitigate random errors introduced by experiments, all results were averaged over multiple trials. Figure 1 illustrates the variation in reconstruction error for Test Case 1 with respect to the number of POD bases and measurement points. Data from 10, 20, 30, 50, 70, and 100 randomly located measurement points were used, and the plotted data represent the average of multiple calculations to eliminate the influence of measurement point locations. As shown in the figure, the reconstruction error for a single POD base consistently exceeds 0.2, significantly higher than the errors observed in the other four cases. As the number of POD bases increases, the reconstruction error does not exhibit substantial variation. Additionally, while the reconstruction error decreases with an increase in the number of measurement points, the reduction is not pronounced. For instance, when the number of POD bases is 2 and the number of measurement points is 10, the reconstruction error is 0.15. Even when the number of measurement points increases to 100, the error only marginally decreases to 0.145. This indicates that a POD basis number of 2, combined with a relatively small number of measurement points, is sufficient to achieve a satisfactory reconstruction of the temperature field.

### 3.2. Sensor Deployment Strategies and Reconstruction Performance

To investigate the impact of different operating conditions (6 water intake depths) and fixed sensor placement (measurement point locations) on the reconstruction results, Figures 2 and 3 depict the reconstruction performance for surface fixed measurement points and vertical fixed measurement points at the dam front under the six operating conditions, respectively. The results indicate that both sparse representation and POD methods are capable of effectively reconstructing the temperature field, with errors generally concentrated around 0.15. Specifically, the temperature field reconstructed using sparse representation exhibits richer and more detailed features, and the overall reconstruction performance is slightly superior to that of the POD method. Notably, under vertical fixed measurement point conditions, the performance of sparse representation improves by 30% compared to POD. These findings underscore the advantages of sparse representation in capturing finer spatial details and achieving higher reconstruction accuracy, particularly in complex scenarios with varying operational conditions and sensor placements.

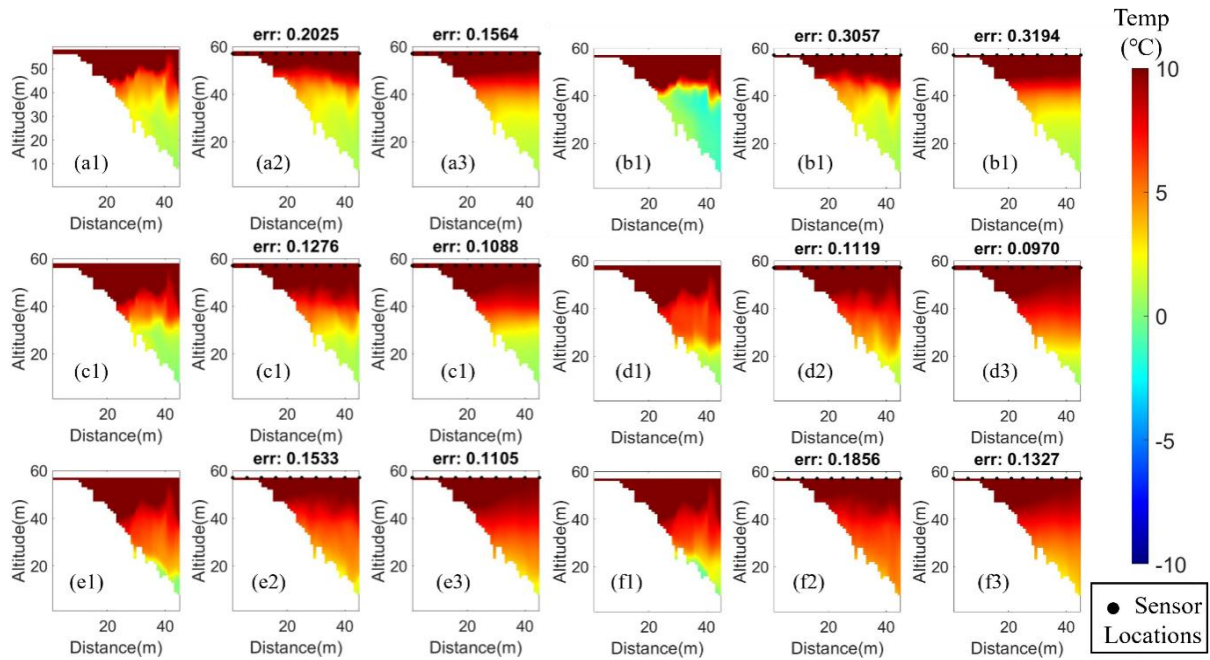

Figure 2 Reconstruction Performance of Fixed Water Surface Measurement Points Under Different Water Intake Depth Conditions(a-f represent different working conditions, i.e., different water intake depths (5, 15, 25, 35, 45, 55 m). 1-3 denote actual the temperature field, the temperature field reconstructed by sparse representation, and the temperature field reconstructed by POD, respectively.)

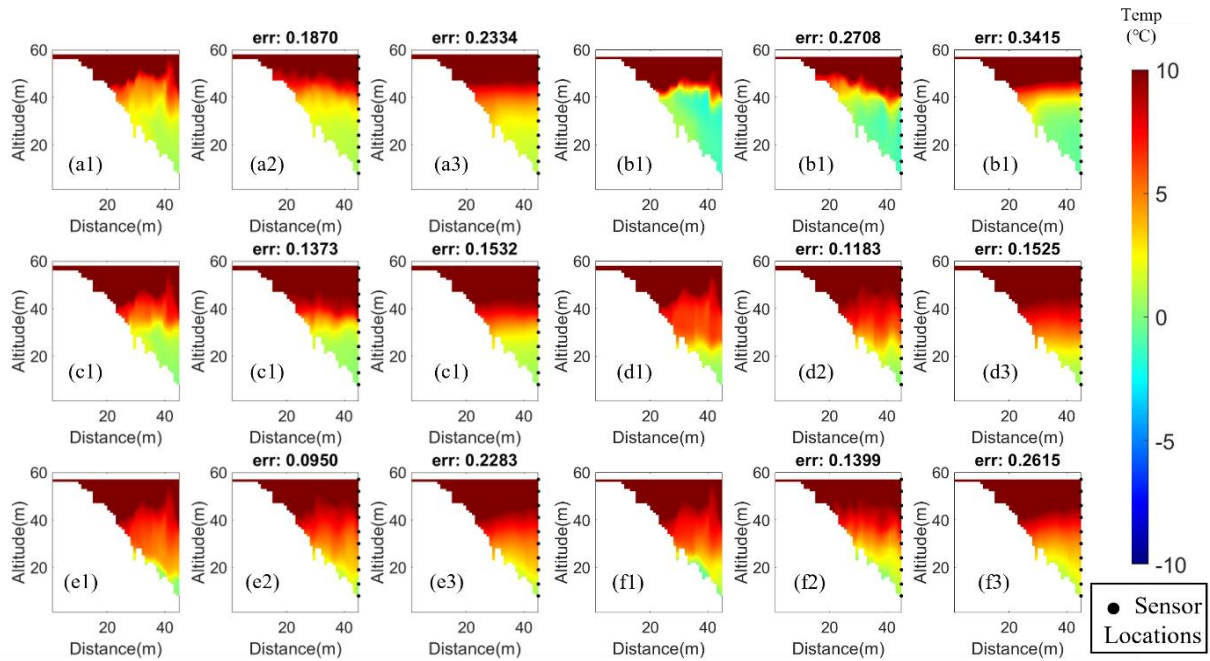

Figure 3 Reconstruction Performance of Vertically Fixed Measurement Points Under Different Water Intake Depths

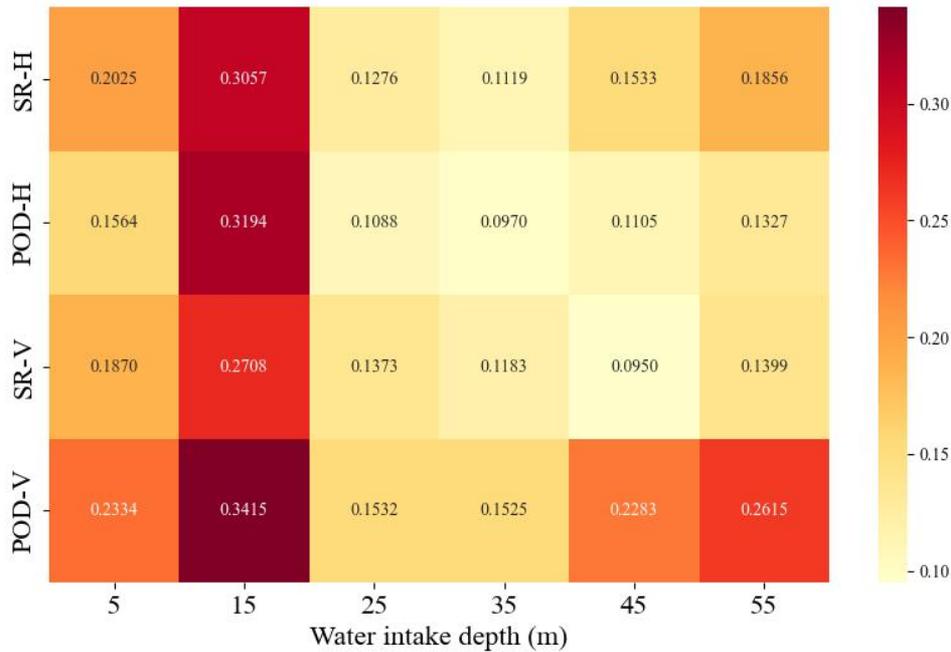

Figure 4 The reconstruction errors of POD and sparse representation at the vertical fixed measurement points below the water surface and in front of the dam.

Figure 4 illustrates the reconstruction errors of sparse representation and POD under two fixed measurement point configurations: surface and vertical placements. It is evident that POD exhibits significant variability in performance between the two measurement point arrangements, with a reconstruction error difference of nearly 50%. In contrast, sparse representation demonstrates superior robustness, with only a 14% difference in reconstruction errors between the two configurations. This enhanced stability aligns with the inherent noise-suppressing properties of L1 regularization in sparse representation, which naturally contributes to its robustness. As a result, sparse representation achieves a more balanced and consistent performance across different measurement point arrangements, further underscoring its reliability in temperature field reconstruction under varying sensor placement scenarios.

### 3.3. Spatial Analysis of Temperature Reconstruction Errors

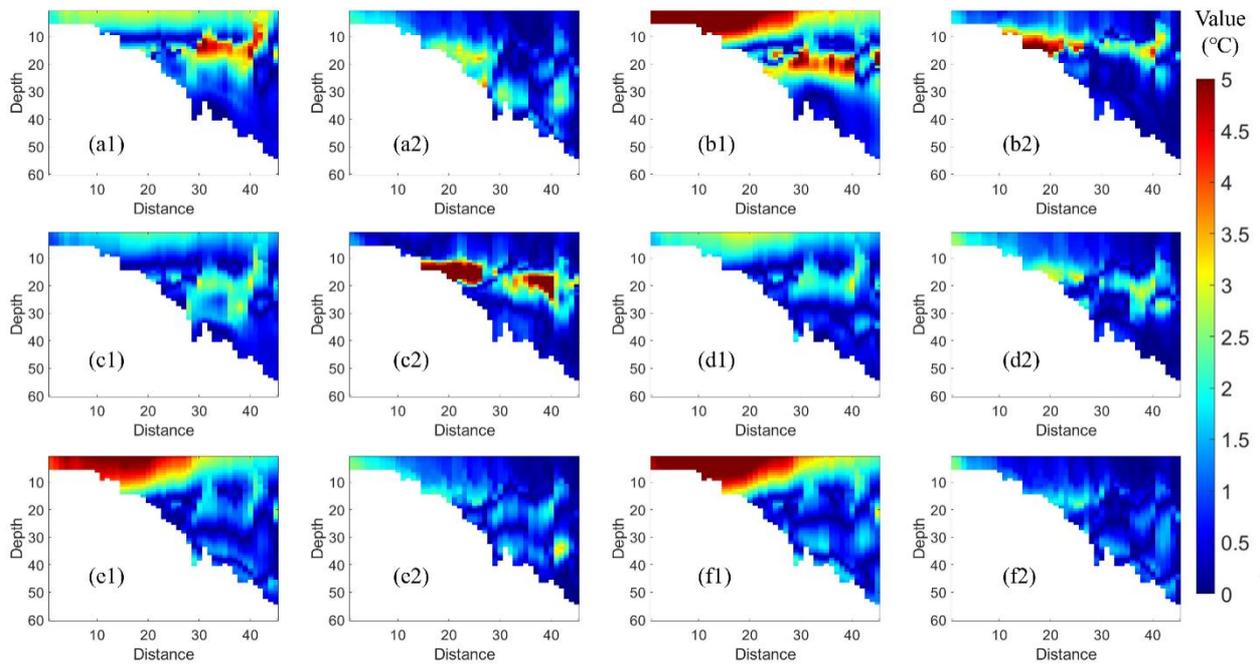

Figure 5 Spatial Distribution of Temperature Field Reconstruction Errors with Fixed Water Surface Measurement Points(a-f represent different working conditions, i.e., different water intake depths (5, 15, 25, 35, 45, 55 m). 1 and 2 denote the error distribution between the temperature field reconstructed by sparse representation and the actual temperature field, and the error distribution between the temperature field reconstructed by POD and the actual temperature field, respectively.)

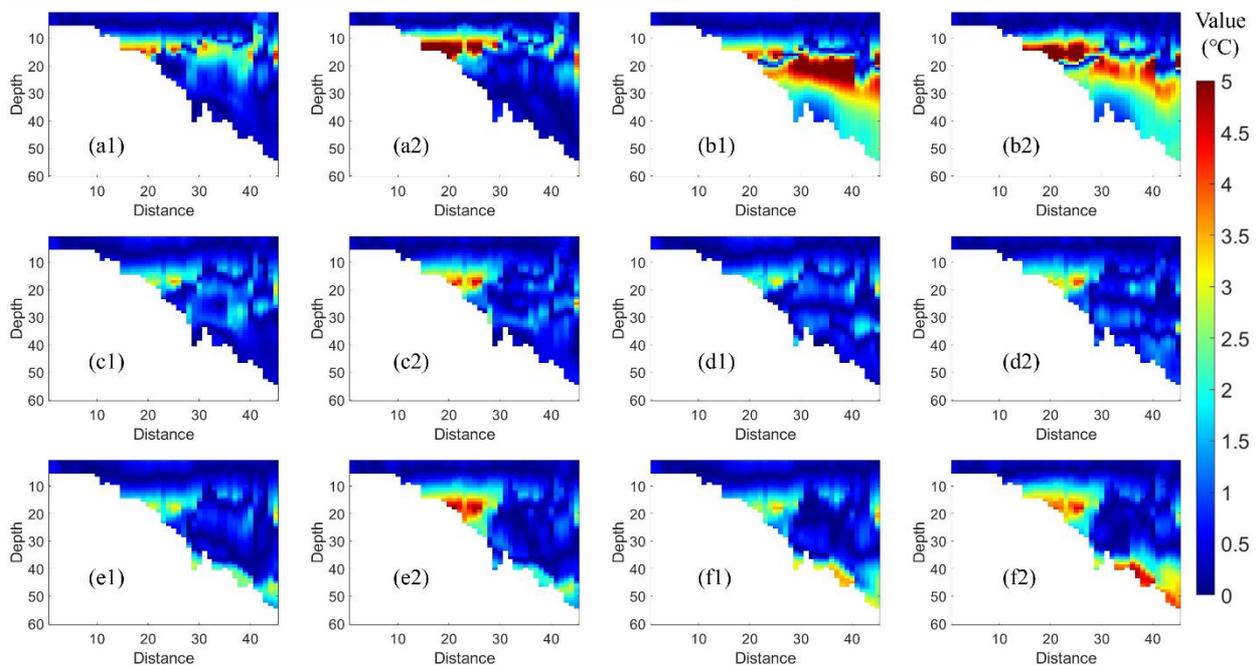

Figure 6 Spatial Distribution of Temperature Field Reconstruction Errors with Vertically Fixed Measurement Points

To further analyze the error performance of POD and sparse representation under fixed measurement point configurations, we consider factors such as the disturbance of the temperature field caused by water intake at different depths and the distance between measurement points and reconstruction points. We examine the spatial distribution of errors for POD and sparse representation under two fixed measurement point arrangements across six water intake depths. Figure 5 illustrates the spatial distribution of temperature field reconstruction errors for POD and sparse representation under the surface fixed measurement point arrangement, while Figure 6 shows the corresponding distribution under the vertical fixed measurement point arrangement at the dam front. Both figures reveal a distinct stratification phenomenon, where the reconstruction errors in the upper layers of the temperature field are consistently lower than those in the lower layers. This observation suggests that the reconstruction accuracy is influenced by the vertical structure of the temperature field, with the upper layers being more accurately reconstructed due to potentially less complex thermal dynamics and closer proximity to the measurement points. These findings provide valuable insights into the spatial variability of reconstruction errors and highlight the importance of considering vertical stratification in temperature field reconstruction.

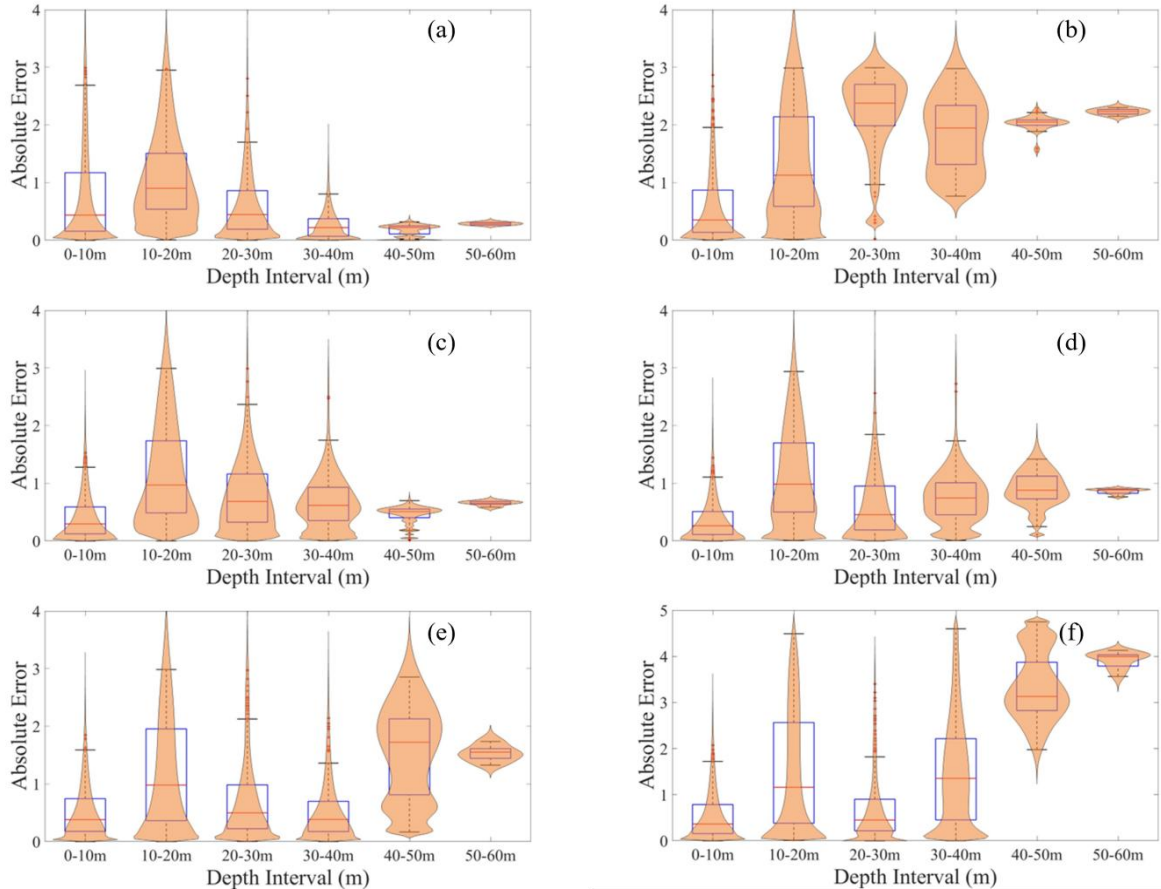

Figure 7 Error Distribution of Temperature Field Reconstructed by Sparse Representation in Different Depth Intervals(a-f represent different working conditions, i.e., different water intake depths (5, 15, 25, 35, 45, 55 m))

Different water intake depths cause fluctuations in water bodies at varying water levels, thereby influencing the temperature field. The temperature field reconstructed using sparse representation also reflects this characteristic. As shown in Figure 6, when the water intake depth is shallow, the errors across different water level intervals remain relatively concentrated. For instance, at a water intake depth of 5 meters, except for the 10-20 meter depth interval, the absolute errors in the other five intervals are clustered around 0.5. However, as the water intake depth increases, the errors across different water level intervals exhibit more pronounced variability, with larger errors observed at deeper water levels. This phenomenon can be attributed to the reservoir's geographical structure, where the water body becomes increasingly narrow with greater depth. Consequently, the number of measurement points in the 40-60 meter depth interval is reduced, which adversely affects the reconstruction performance of sparse representation.

Similarly, in Figure 8, the reconstruction results using POD also exhibit analogous trends and distributions. The errors in the temperature field reconstruction show a clear stratification, with smaller errors in the upper water layers and larger errors in the deeper layers. This pattern aligns with the observations made in the sparse representation results, further confirming the influence of water intake depth and reservoir geometry on the reconstruction accuracy. The consistency between the two methods underscores the challenges posed by deeper and more confined regions in the reservoir, where the reduced number of measurement points and complex thermal dynamics contribute to increased reconstruction errors. These findings reinforce the importance of considering vertical stratification and spatial variability when reconstructing temperature fields in complex aquatic environments.

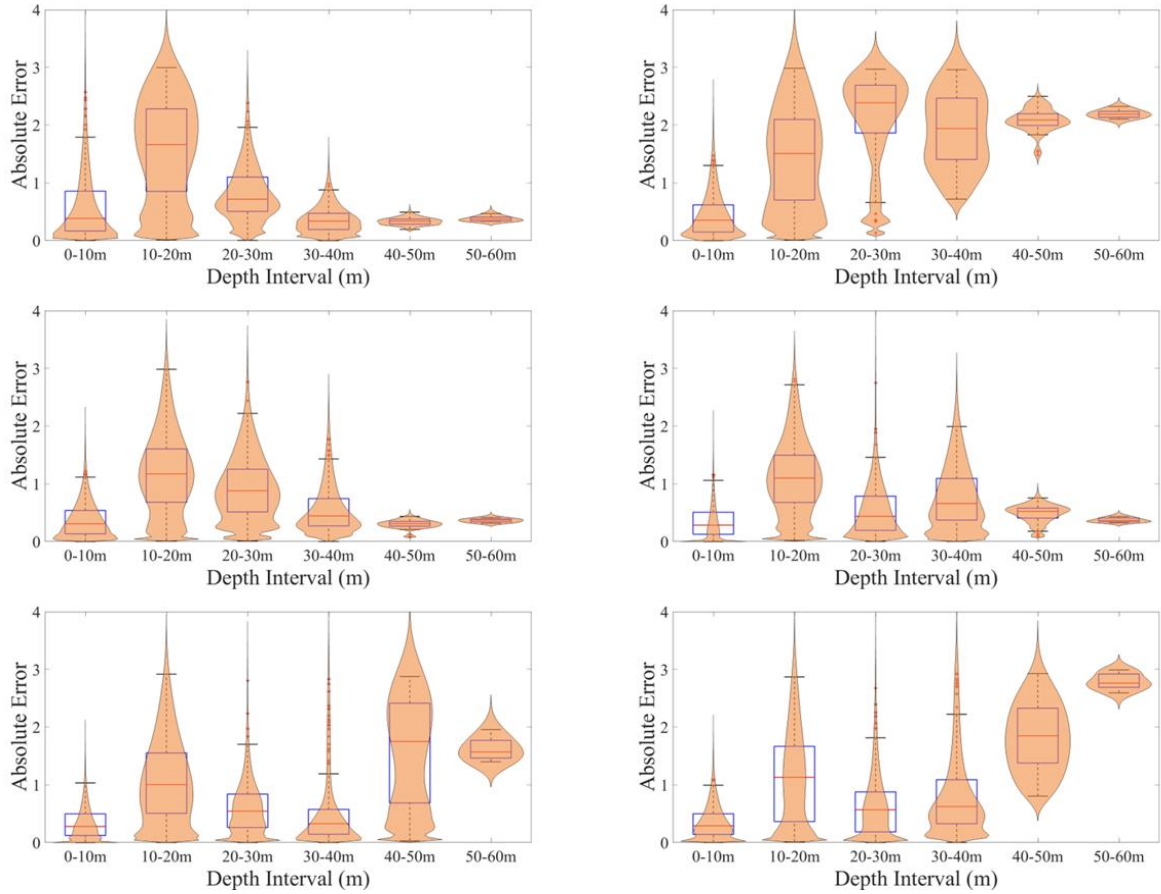

Figure 8 Error Distribution of Temperature Field Reconstructed by POD in Different Depth Intervals

## 4. Conclusion

In this study, we employed two flow field reconstruction methods, POD and sparse representation, to reconstruct the temperature field of Lake Diefenbaker Reservoir in Saskatchewan, Canada. For the reconstruction of the temperature field based on limited known measurement data, Gappy POD was utilized to achieve rapid and accurate reconstruction using a small number of POD bases. The reconstruction error decreases as the number of POD bases and measurement points increases, eventually stabilizing. This demonstrates that a small number of POD bases can effectively reconstruct the temperature field from limited local measurement data.

Next, we investigated the reconstruction performance of POD and sparse representation under six operating conditions (water intake depths of 5 m, 15 m, 25 m, 35 m, 45 m, and 55 m) for two fixed measurement point configurations: surface placement and vertical placement at the dam front. The results indicate that both POD and sparse representation can reconstruct the temperature field well in most cases. Due to the L1 regularization in sparse representation, it is less sensitive to the distribution of measurement points, resulting in more consistent performance across the two configurations. In contrast, POD is significantly influenced by the distribution of measurement points, with a performance gap of nearly 50% between the two configurations. This suggests that sparse representation is more suitable for handling complex and nonlinear problems.

Furthermore, we analyzed the spatial distribution of reconstruction errors in the temperature field. Sparse representation exhibits richer details, which aligns with its inherent robustness. However, both methods perform poorly in the lower water layers, particularly as the water intake depth increases. This is because the lower water layers experience greater fluctuations compared to the upper layers, and the energy contribution from the lower layers is relatively low. Additionally, the sparse distribution of measurement points in the lower layers leads to insufficient data support, further exacerbating the reconstruction challenges. These findings highlight the need for further exploration of optimal measurement point placement strategies to improve reconstruction accuracy, especially in deeper and more dynamic regions of the reservoir.